\documentclass[10pt,twocolumn]{article} 
\usepackage{simpleConference}
\usepackage{times}
\usepackage{graphicx}

\usepackage{amsmath,amssymb} 
\usepackage{color}
\usepackage{siunitx}
\usepackage[table]{xcolor}
\usepackage{caption}
\usepackage{epsfig} 
\usepackage{array}
\usepackage{booktabs}
\usepackage{multirow}
\newcommand{\head}[1]{\textnormal{\textbf{#1}}}

\begin{document}

\title{UnDEMoN 2.0: Improved Depth and Ego Motion Estimation through Deep Image Sampling}

\author{Madhu Babu V,  Swagat Kumar, Anima Majumder and Kaushik Das  \\
\\
Technical Report \\
TATA Consultancy Services, Bangalore, India. \\
\today
\\
\\
 (madhu.vankadari,  swagat.kumar, anima.majumder, kaushik.da)@tcs.com  \\
}

\maketitle
\thispagestyle{empty}

\begin{abstract}
In this paper, we provide an improved version of UnDEMoN model for depth and ego motion estimation from monocular images. The improvement is achieved by combining the standard bi-linear sampler with a deep network based image sampling model (DIS-NET) to provide better image reconstruction capabilities on which the depth estimation accuracy depends in un-supervised learning models. While DIS-NET provides higher order regression and larger input search space, the bi-linear sampler provides geometric constraints necessary for reducing the size of the solution space for an ill-posed problem of this kind. This combination is shown to provide significant improvement in depth and pose estimation accuracy outperforming all existing state-of-the-art methods in this category. In addition, the modified network uses far less number of tunable parameters making it one of the lightest deep network model for depth estimation. The proposed model is labeled as "UnDEMoN 2.0" indicating an improvement over the existing UnDEMoN model. The efficacy of the proposed model is demonstrated through rigorous experimental analysis on the standard KITTI dataset. 
\end{abstract}
\section{Introduction}

%
Depth and Ego motion estimation from images is an important problem in computer vision which finds application in several fields such as augmented reality \cite{pose_ar} \cite{taylor2016efficient}, 3D construction \cite{stereoscan}, self-driving cars  \cite{handa2014benchmark}, medical imaging \cite{mahmood2018deep} etc. Recent advances in deep learning have helped in achieving new benchmarks in this field which is getting better and better with time. The initial deep models \cite{eigen2014depth} \cite{learning_depth_mono_cnf} used supervised mode of learning that required explicit availability of ground truth depth which is not always possible. This is partially remedied by using semi-supervised methods which either use sparse ground truth obtained from sensors like LIDAR \cite{kuznietsov2017semi} or make use of synthetically generated data as ground truth \cite{luo2018single}. Compared to these methods, the unsupervised methods are becoming more popular with time as no explicit ground truth information is required for the learning process. In these cases, the geometric constraints between a pair of images either in temporal \cite{zhou2017unsupervised} \cite{mahjourian2018unsupervised} or spatial domain \cite{monodepth17} or both \cite{babu2018deeper} are exploited to estimate the depth and pose information. Continuing this trend, we restrict our focus of discussion only to unsupervised deep network models from here onwards in the rest of this paper. Some of the most recent and best results in this category are reported by methods such as, Vid2Depth \cite{mahjourian2018unsupervised}, UnDeepVO \cite{li2018undeepvo}, DeepFeat-VO \cite{zhan2018unsupervised} and UnDEMoN \cite{babu2018deeper}. Vid2Depth \cite{mahjourian2018unsupervised} uses inferred 3D world geometry and enforces consistency of estimated point clouds and pose information across consecutive frames. Since they rely on temporal consistency (monocular sequence of images), the absolute scale information is lost. This is remedied in UnDeepVO \cite{li2018undeepvo} where authors enforce both spatial and temporal consistencies between images as well as between 3D point clouds. UnDEMoN \cite{babu2018deeper} further improves the performance of UnDeepVO \cite{li2018undeepvo} by predicting disparity instead of depth and using different penalty function for training. DeepFeat-VO \cite{zhan2018unsupervised} attempts to further improve the results by including deep feature-based warping losses into the training process. These deep features are obtained from a depth model pre-trained on a different dataset. 

In spite of these advancements, the depth and pose estimation results are still not close to what is obtained from stereo methods \cite{luo2018single} which use left-right image pair as input to the network. There is still enough scope for improving the accuracies of unsupervised methods.  Most of the unsupervised methods try to minimize the reconstruction losses either in temporal domain (forward-backward images) \cite{zhou2017unsupervised} \cite{mahjourian2018unsupervised} or in spatial domain (left-right images) \cite{monodepth17} or in both \cite{babu2018deeper} during the training process. Bi-linear interpolation is one of the commonly used method for image reconstruction which is widely used in the view synthesis literature \cite{monodepth17} \cite{zhou2017unsupervised}. Bi-linear interpolation uses a quadratic regression model to estimate color intensity of a target pixel by using only four neighboring pixels of the source image. The use of only four neighborhood points limits the ability of the regression model to deal with large motions in the scene. This is remedied by using a large search space (more than four points) to estimate the target pixel as suggested in \cite{mahjourian2018unsupervised}. This, however, may not give good interpolation for far-away objects. Furthermore, the accuracy of reconstruction can be improved by using higher order regression model at the cost of higher computational cost. This provides us the necessary motivation to use a deep network model for image reconstruction required for disparity estimation. This deep network provides higher order regression which is expected to perform better a quadratic regressor used in bi-linear sampler. In addition, it solves the problem of search space required for interpolation as it takes the whole image as the input. The use of deep network based image sampler instead of using the standard bi-linear sampler commonly used in other methods provides the basis for the work presented in this paper. 

\begin{figure}[!t]
\centering
\label{network}
\includegraphics[scale= 0.3]{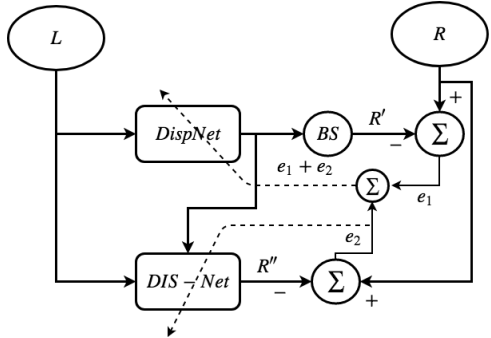} 
\caption{\small Architecture of UnDEMoN 2.0. It uses a deep network based image sampler (DISNet) combined with a bi-linear image sampler (BIS) to provide superior image reconstruction capabilities required for training the depth estimation network (DispNet).}
\label{fig:net_arch}
\end{figure}

In this paper, we suggest several improvements to the UnDEMoN architecture \cite{babu2018deeper} with the aim to produce superior depth and pose estimation results. The proposed model makes use of a deep network for reconstructing right image given the left image and the corresponding disparity. This is labeled as deep image sampler network (DISNet) which is used along with a bi-linear sampler (BIS) to provide superior image reconstruction abilities required for depth and pose estimation. While the DISNet provides higher order regression and larger input search space for interpolation, the bi-linear sampler provides the necessary geometry constraints to reduce the solution space of this ill-posed problem. A simplified overview of the proposed framework is shown in Figure \ref{fig:net_arch}. As one can observe, the disparity estimation network (DispNet) uses the image reconstruction error obtained from both DISNet and BIS together for training.  In addition to this, we use a thinner deep model (with lesser number of hidden layers) compared to the previous UnDeMoN model. The total number of parameters in the proposed inference model is about one-fourth (only 25\%) of the original UnDEMoN model \cite{babu2018deeper}, making it one of the lightest depth or disparity estimation model in the literature. The resulting model provides superior depth and pose estimation which outperforms all existing state-of-the-art methods. For instance, the proposed model provides 4.47\%, 14\%, 8.86\% and 18\% improvement over UnDEMoN \cite{babu2018deeper}, Vid2Depth \cite{mahjourian2018unsupervised}, DeepFeat-VO \cite{zhan2018unsupervised} and UnDeepVO \cite{li2018undeepvo} respectively. The proposed model is named "UnDEMoN 2.0" to indicate improvement over the previously existing UnDEMoN model \cite{babu2018deeper}.

In short, the following two main contributions are made in this paper: (1) The proposed model makes use of a deep network based image sampling model for depth and pose estimation which is a novel contribution in this field. (2) The proposed model uses very less number of trainable parameters for achieving superior performance making it one of the most computationally efficient deep model for depth and pose estimation. our model. Finally, a rigorous analysis of the network performance is carried out on KITTI dataset and is compared with all existing state-of-the-art methods. To the best of our knowledge, the results presented in this paper are the best reported so far and hence forms a new benchmark in this field. 

The rest of this paper is organized as follows. An overview of related work is provided in the next section. The proposed model is described in Section \ref{sec:meth}. The experimental results are analysed and compared in Section \ref{sec:expt}. Finally, the conclusion and future scope of improvement is provided in Section \ref{sec:conc}. 
\section{Related Work}

Depth and ego-motion estimation from images has a long history in the literature. Initially researchers were using conventional techniques to solve this problem. Some of those include, stereo and feature matching techniques \cite{scharstein2002taxonomy}. It is until recently, with the advancement of deep learning techniques and up-gradation of high performance computational devices, like GPU machines, the computer vision researchers started applying deep learning techniques \cite{furukawa2015multi, demon, disp_net, monodepth17} and have shown significant improvements over the conventional approaches. 
 All these deep learning techniques can be broadly categorized as supervised (using ground-truth depth data), semi-supervised and unsupervised approaches. Most of the pioneering works in this direction learn depth from images in a supervised manner by using ground-truth data from depth sensors \cite{demon, eigen2014depth, learning_depth_mono_cnf, disp_net, kendall2015posenet, deepvo}. However, over the period of time, the researches have inclined towards the unsupervised mode of solving the problem, mainly due to the constraint of generating ground-truth depth data needed to train the network. So far, very few works have been reported towards unsupervised depth and pose estimation by using deep neural networks. Among those few techniques, Garg et al. \cite{garg2016unsupervised} first introduced the concept of single-view depth estimation by using calibrated stereo images. This approach is soon adopted and few works with promising results are followed in the subsequent years. Some of those techniques include \cite{monodepth17, zhou2017unsupervised, babu2018deeper, mahjourian2018unsupervised}. Godard et al. \cite{monodepth17} introduced a loss function that incorporates the consistency between the disparities obtained from both the left and right pair images and thereby beating the then state-of-the art techniques on KITTI driving database. However, the approach needed both left and right images during the training process. In an attempt to overcome this limitation of using both the left and right pair images during training, Zhou et al. \cite{zhou2017unsupervised} first introduced a concept of using only $n$ temporally aligned snippets of images to train the network. The work jointly estimates depth and camera pose by using temporal reconstruction loss.  Nevertheless, one major concern with this technique is that, the absolute scale information in depth prediction goes missing on using monocular images alone. This is further remedied by Li et al. \cite{li2018undeepvo}, where they combine both the spatial and the temporal reconstruction losses to jointly predict scale-aware depth and pose directly from monocular stereo images. Inspired by this approach, Babu et al. \cite{babu2018deeper} have further extended the work and showed promising depth and pose estimation results on the KITTI dataset. They have incorporated a differential variant of absolute norm named Charbonnier penalty \cite{charbonnier_loss} to the objective function. 

In most of the aforesaid approaches, bi-linear interpolation technique is applied for image warping. Although, the method is readily adopted by many of the researchers due to its simpler mathematical representation and lesser computational cost, it has some important limitations, specially while estimating depth and pose \cite{wang2018occlusion}. Bi-linear interpolation uses only four neighborhood pixels, thereby limiting the use of a larger search space needed for large/fast object motion. This has been addressed in \cite{wang2018occlusion}, where the authors introduced a novel image warping technique with larger search space to get better flow prediction for objects with larger motion. This approach partially solves the problem, in a sense, that the use of a larger search space for far away objects will introduce unnecessary extract features, that will intern deteriorate the interpolation accuracy. Another work in this direction is proposed by Mahjourian et al. \cite{mahjourian2018unsupervised}. Instead of using the bi-linear interpolation for image warping they have used inferred 3D geometry of the entire image to enforce consistency of the estimated
3D point clouds and ego-motion across consecutive frames. 

In contrast, we introduce a deep learning network, named \emph{DIS-net} along-with the bi-linear interpolation module to jointly participate in image warping, thereby improving the depth and pose estimation accuracy. 

\section{The Proposed Method} \label{sec:meth}

In this section, we provide the details of modifications that is applied to the UnDEMoN architecture \cite{babu2018deeper} with an aim to improve its performance. The UnDEMoN architecture is composed of two deep networks namely, DispNet and PoseNet. DispNet estimates disparity directly from a monocular left image which is then used for estimating absolute scale aware depth. On the other hand, the Pose Net estimates the ego motion from a sequence of temporarily aligned monocular left images (\emph{snippets}). A bi-linear sampler is used as a spatial image warping module to predict right images given the left image and the left-to-right disparity.  The overall depth and pose estimation accuracy is improved by training the combined model with a Charbonier Penalty function that incorporates both temporal and spatial reconstruction losses. 

The modifications carried out to the UnDEMoN architecture is shown in Figure \ref{fig:net_arch}. The modifications include using a deep network based image sampler called DISNet along with the standard bi-linear sampler (BIS) to compute image reconstruction losses necessary for training the DispNet. In this figure, $L$ and $R$ refers to the left and right images of the stereo pair. Similarly $R'$ and $R''$ refer to the reconstructed images obtained using bi-linear sampler (BIS) and DISNet respectively. The DISPNet is trained using the combined reconstruction losses obtained from these two networks. In addition, a thinner version of encoder-decoder model used in UnDEMoN architecture is used to reduce the total number of tunable parameters thereby reducing the overall computational complexity and memory footprint of the deep model. This new architecture is labeled as "UnDEMoN 2.0" indicating an improvement over the existing model. The details of DISNet and necessary motivations behind this are described below in this section.

\begin{figure}[!t]
\centering
\includegraphics[height=4.5cm]{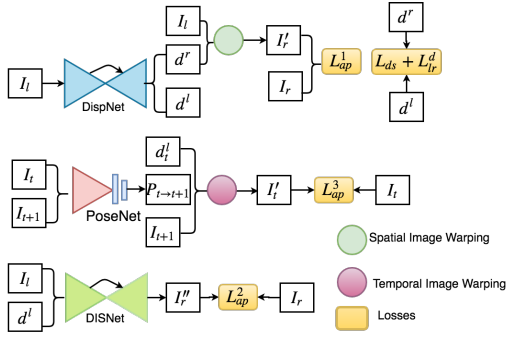} 
\caption{\small Detailed overview of UnDEMoN 2.0 architecture. It uses three deep network models, namely, DispNet, PoseNet and DISNet. DISNet is a deep network based image sampler that provides better image reconstruction capabilities compared to a standard bi-linear sampler.}
\label{fig:net_arch2}
\end{figure}

\subsection{Image Warping using DISNet}

While the Figure \ref{fig:net_arch} provides a simplified overview of the proposed architecture, the details of underlying processes are shown in Figure \ref{fig:net_arch2}.  The DispNet module takes the left image $I_l$ as input and gives left-to-right disparity $d^{l}$ and right-to-left disparity $d^{r}$ as output. The bi-linear sampler (BIS) based spatial image warping module uses $I_l$ and $d^{r}$ to reconstruct the right image denoted by $I'_r$. By using the right image as the ground truth $I_r$, the spatial image reconstruction loss for this module is denoted by the symbol $L^1_{ap}$ which is also known as the image appearance loss. Similarly, disparities $d^{r}$ and $d^{l}$ are used for computing the smoothness loss $L_{ds}$ and disparity left-right consistency loss $L^d_{lr}$.  The PoseNet takes a temporally aligned snippet of $n$ monocular images (for $n=2$, snippet is $\{I_t, I_{t+1}\}$) as input and gives ego motion between the images $P_{t\rightarrow  t+1}$. Another bi-linear sampler based temporal image warping module uses $d_t^{l}$, $I_{t+1}$ and $P_{t\rightarrow t+1}$ to reconstruct the temporally aligned image $I'_t$. Taking $I_t$ as the input, the temporal image reconstruction loss $L^3_{ap}$ is computed. The DISNet takes $I_l$ and $d^{r}$ as input and reconstructs the right image denoted by $I''_r$. This together with $I_r$ as ground truth gives rise to appearance loss denoted by $L^2_{ap}$. All these losses are used for tuning the weight parameters for all the three networks simultaneously. These training losses are described next in this section.

\subsection{Training Losses}

\subsubsection*{Appearance loss}
The appearance loss represents the error between the reconstructed image and the actual image. It is mathematically represented as: 
\begin{equation}
 \label{eq:aplos}	
 L_{ap} =\frac{1}{N}\sum_{ij}{\alpha}\rho(1-SSIM(I_{ij},\tilde{I}_{ij})+ (1-\alpha)\rho\|I_{ij}-\tilde{I}_{ij}\|
\end{equation}
where $I$ and $\tilde{I}$ are the original and reconstructed images and $\|.\|$ represents the $L_1$ error norm between these two. The SSIM is the structural similarity index \cite{wang2004image} and $\rho$ is the Charbonier penalty function \cite{babu2018deeper}. The parameter $\alpha < 1$ defines the weight given to the $L_1$ norm loss and the SSIM to form a convex combination.


\subsubsection*{Disparity Smoothness:}
\label{disp_loss_sec}
We encourage the disparity to be smooth locally by taking the $L_1$ norm of disparity gradients $\partial d$. As depth discontinuities often occurs at the image gradients, we use an edge aware term using image gradients $\partial I$ to weight the disparity gradients $\partial d$. Mathematically it can be defined as :
\begin{equation}
 \label{gradient_loss}
 L_{ds}=\frac{1}{N}\sum_{ij}\rho({\partial_x d_{ij}e^{-||\partial_xI_{ij}||}})+
			    \rho({\partial_y d_{ij}e^{-||\partial_yI_{ij}||}})
\end{equation}

\subsubsection*{Left-Right Consistency:}

\label{lr_consistancy}
 To further improve the coherence between the estimated disparities, we project the disparities from one to another and take the $L_1$ norm between them. In a way, this helps in enforcing the cycle consistency between the predicted disparities \cite{monodepth17}. For instance, consistency loss with left-to-right disparity can be defined as: 
\begin{equation}
 \label{lr_loss}
 L^{d_l}_{lr}=\frac{1}{N}\sum_{ij}|d^{l}_{ij}-d^{r}_{ij+d^{l}_{ij}}|
\end{equation}
 
The total loss $L$ includes appearance losses calculated using left and right images from all the three networks ($L^1_{ap}$, $L^2_{ap}$,$L^3_{ap}$), disparity smoothness loss $L_{ds}$ and left-right consistency loss $L^d_{lr}$. These are given as follows:
\begin{eqnarray}
 L^1_{ap} &=& \lambda^1_{ap}[L^l_{ap}(I_l,I'_l)+L^r_{ap}(I_r,I'_r)] \nonumber \\
 L^3_{ap} &=& \lambda^3_{ap} L^{t+1}_{ap} (I_t, I'_t) \nonumber \\
 L^2_{ap} &=& \lambda^2_{ap}[L^l_{ap}(I_l,I''_l)+L^{r}_{ap}(I_r,I''_r)] \nonumber \\
 L_{ds} &=&  \lambda_{ds}(L^l_{ds}+L^r_{ds}) \nonumber \\
 L^d_{lr} &=& \lambda^d_{lr}(L^{d_l}_{lr}+L^{d_r}_{lr}) \nonumber \\
 L &=& L^1_{ap}+L^2_{ap}+L^3_{ap}+L_{ds}+L^d_{lr}
  \label{eq:losses}
\end{eqnarray}

\subsection{Effect of Image Reconstruction using DISNet}
\begin{figure*}[!t]
\centering
\includegraphics[scale= 0.7]{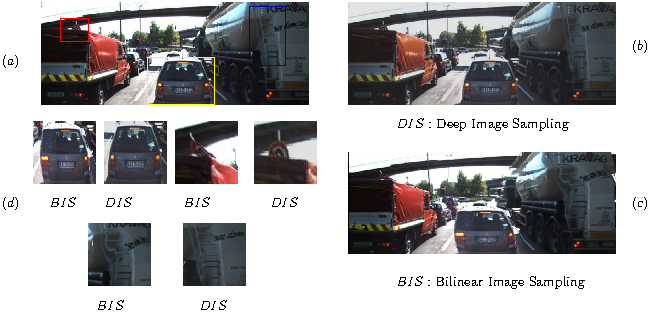} 
\caption{\small Effect of Deep Image Sampling on image reconstruction. As one can observe, DISNet provides better reconstruction compared to the conventional bi-linear sampler. (a) is the original image, (b) is the image reconstructed using DISNet, (c) is the image reconstructed using bi-linear sampler (BIS) and (d) shows the crop outs which highlights the difference in reconstruction for both the methods.}
\label{fig:bisdis}
\end{figure*}

\begin{figure}[!t]
\centering
\includegraphics[height=5cm]{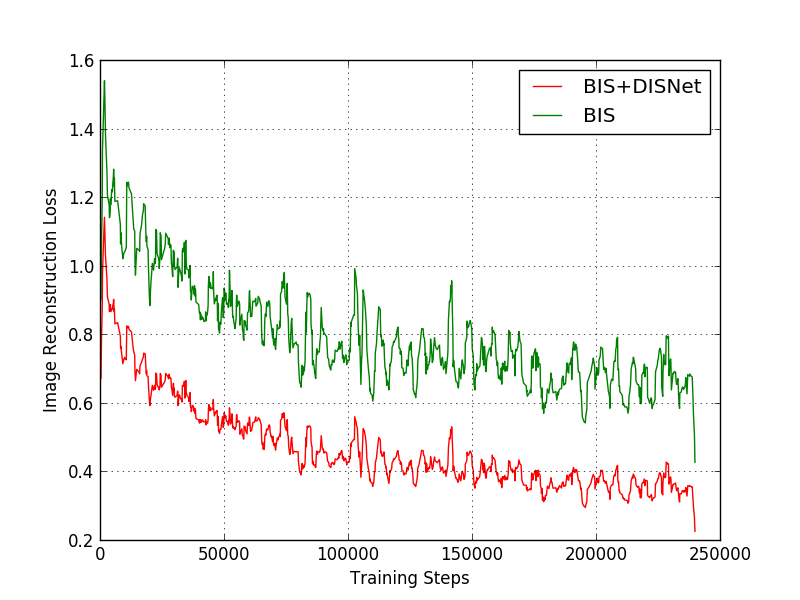} 
\caption{\small The plot of image reconstruction loss during training. The combination of DISNet and BIS leads to lower reconstruction losses compared to models that only use bi-linear sampler (BIS).}
\label{fig:reconst_loss}
\end{figure}

The effect of Deep Image Sampler on the image reconstruction is shown in Figure \ref{fig:bisdis}. In this figure (a) shows the original image, (b) and (c) are the reconstructed images obtained using DISNet and bi-linear sampler (BIS) respectively. As one can observe, BIS provides better reconstruction for low frequency information such color but fails to reconstruction the structural attributes. In comparison, DISNet provides better structural reconstruction compared to bi-linear sampler at the cost of low frequency information. Hence the combination of these two is expected to provide better reconstruction performance. The cropped out regions high-lighting the differences between BIS and DISNet reconstructions are shown in Figure \ref{fig:bisdis} (d). This observation is further confirmed by the plots shown in Figure \ref{fig:reconst_loss} where one can clearly see that the combination of DISNet and BIS provides better lower reconstruction losses compared to the models that only use bi-linear sampler for image reconstruction.  The effect of DISNet on the overall depth and pose estimation accuracy will be described next in the experimental section. 

\begin{figure*}[tphb]
\centering
\includegraphics[scale=0.17]{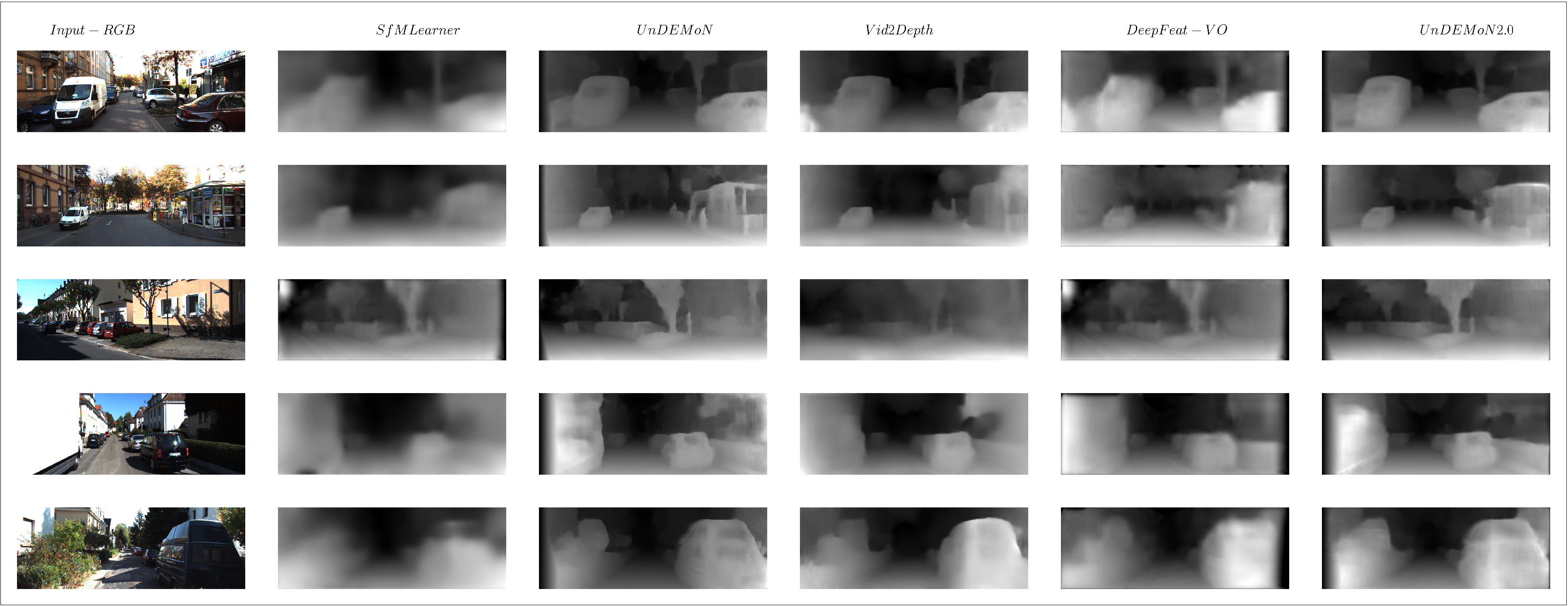}
\caption{\small The UnDEMoN 2.0 results on KITTI Eigen split \cite{eigen2014depth} dataset compared to the state-of-the-art methods like  SfMLearner \cite{zhou2017unsupervised},UnDEMoN \cite{babu2018deeper} and DeepFeat-VO \cite{zhan2018unsupervised}. As one can observe, our proposed method provides better depth estimate compared to these methods. }
\label{fig:depth_results}
\end{figure*}

\begin{table*}[!t]
\caption{\small Performance Comparison of UnDEMoN 2.0 with existing state-of-the-art techniques using Eigen split \cite{geiger2013vision}. Results for Liu et al. \cite{liu2015deep} are taken from \cite{monodepth17}. The Eigen et.al results are recomputed with Velodyne laser data. For fair comparison, the Eigen and Garg results are computed according to crop described in \cite{eigen2014depth} and \cite{garg2016unsupervised}. The first part of the Table show results of Eigen split with Garg crop with 80 meter maximum depth. Similarly, second part show the results for Garg crop with 50 meters of maximum depth.
The column Supervision refers D as Depth, M as Monocular, MS as Monocular Stereo. These are the supervisions used while training. Our rows refer D as only depth network, D+DIS as the combination of Depth+DISNet and UnDEMoN 2.0 as the combination of Depth + DIS + PoseNet. The cells in blue color show the accuracy metric (higher is better) and the remaining columns give error metrics (lower is better).}
\label{tab:kitti_eigen_result}
\centering
\scriptsize{
\renewcommand{\arraystretch}{1.2}
\begin{tabular}{ |p{1.9cm}|p{1cm}||p{1cm}|p{1cm}|p{1cm}|p{1cm}|p{1.3cm}|p{1.2cm}|p{1.2cm}|p{1.2cm}|p{1.2cm}|  }
 \hline
  Method & Supervision & Abs Rel & Sq Rel & RMSE & logRMSE & \cellcolor[HTML]{7BABF7} $\delta <$1.25
  &\cellcolor[HTML]{7BABF7}$\delta <$1.25$^2$ & \cellcolor[HTML]{7BABF7}$\delta <$1.25$^3$ \\
 \hline
 Train set mean                         & D & 0.361  & 4.826 & 8.102 & 0.377 & 0.638 & 0.804& 0.894 \\ 
 
 Eigen et al.\cite{eigen2014depth} Fine & D & 0.203  & 1.548  & 6.307 & 0.282 & 0.702 & 0.890 & 0.958 \\
 
 Liu et al.\cite{liu2015deep}           & D & 0.201  & 1.584  & 6.471 & 0.273 & 0.680 & 0.898 & 0.967 \\

 Garg et al.\cite{garg2016unsupervised} & MS & 0.152  & 1.226  & 5.849 & 0.246 & 0.784 & 0.921 & 0.967 \\ 

 SfM Learner \cite{zhou2017unsupervised}& M & 0.208  & 1.768  & 6.856 & 0.283 & 0.678 & 0.885 & 0.957 \\ 
 
 Vid2Depth\cite{mahjourian2018unsupervised} & M & 0.163 & 1.240 & 6.220 & 0.250 & 0.762 & 0.916 & 968\\
  
 Monodepth \cite{monodepth17}           & MS & 0.148 & 1.344  & 5.927 & 0.247 & 0.803 & 0.922 & 0.964 \\
 
 UnDeepVO \cite{li2018undeepvo}         & MS & 0.183 & 1.73   & 6.57  & 0.283 &  - & - & - \\
 
 DeepFeat-V0 \cite{zhan2018unsupervised} & MS & 0.144 & 1.391  & 5.869 & 0.241 & 0.803 & 0.928 & 0.969 \\
  
 UnDEMoN \cite{babu2018deeper}          & MS & 0.139 & 1.174  & 5.59  & 0.239 & 0.812 & 0.930 & 0.968 \\
 
 \textbf{Ours D}                        & MS & 0.1365 & 1.1391 & 5.642 & 0.239 & 0.813 & 0.928 & 0.967 \\
 
 \textbf{Ours D+ DIS}                    & MS & 0.1339 & 1.105  & 5.448 & 0.232 & 0.815 & 0.930 & 0.970 \\

 \textbf{UnDEMoN 2.0}               & MS & \textbf{0.1277} & \textbf{1.0125} & \textbf{5.349} & \textbf{0.227} & \textbf{0.823} & \textbf{0.932} & \textbf{0.971} \\
		      
 \hline
 \hline
  
 Garg et al.\cite{garg2016unsupervised} & MS & 0.169  & 1.080  & 5.104 & 0.273 & 0.740 & 0.904 & 0.962\\
 
 Monodepth \cite{monodepth17}           & MS & 0.140  & 0.976  & 4.471 & 0.232 & 0.818 & 0.931 & 0.969 \\
 
 SfM Learner \cite{zhou2017unsupervised}& M  & 0.201  & 1.391  & 5.181 & 0.264 & 0.696 & 0.900 & 0.966 \\
 
 Vid2Depth\cite{mahjourian2018unsupervised} & M & 0.155 & 0.927 & 4.549 & 0.231 & 0.781 & 0.931 & 0.975\\

 DeepFeat-VO \cite{zhan2018unsupervised} & MS & 0.135  & 0.905  & 4.366 & 0.225 & 0.818 & 0.937 & 0.973\\
  
 UnDEMoN \cite{babu2018deeper}          & MS & 0.132  & 0.884  & 4.290 & 0.226 & 0.827 & 0.937 & 0.972 \\
 
 \textbf{Ours D}                        & MS & 0.129  & 0.8344 & 4.259 & 0.225 &  0.827 & 0.935 & 0.972\\
 
 \textbf{Ours D+ DIS}                    & MS & 0.127  & 0.8016 & 4.161 & 0.220 &  0.829 & 0.938 & 0.974\\

 \textbf{UnDEMoN 2.0}               & MS & \textbf{0.121}  & \textbf{0.7619} & \textbf{4.078} & \textbf{0.215} &  \textbf{0.837} & \textbf{0.940} & \textbf{0.974} \\
 
 \hline 
\end{tabular}}
\end{table*}

\section{Experiments \& Results} \label{sec:expt}

This section provides the details of various experiments carried out and the analysis of results obtained from these experiments to validate and corroborate the observations and claims made in the previous section. The deep network architecture shown in Figure \ref{fig:net_arch2} is implemented using TensorFlow \cite{abadi2016tensorflow} which is trained and evaluated using the popular KITTI dataset \cite{geiger2013vision}. The DispNet is a fully convolutional architecture composed of an encoder (from cnv1 to cnv5b) and a decoder (from upcnv5b) with multi scale disparity outputs. We have removed the cnv6, cnv7, upcnv7 and upcnv6 layers of original architecture proposed in UnDEMoN \cite{babu2018deeper} to make it lighter. The decoders uses skip connections \cite{RFB15a} from encoder to improve the quality of the predictions. The PoseNet has a convolutional encoder with two fully connected layers. The DISNet is also similar to the DispNet in architecture except in the prediction layers. In all these networks, Relu \cite{nair2010rectified} activation functions are used for non-linearities except in the output layers. Please refer to the supplementary material for a detailed description about network architectures.  The model has about 19 million parameters and takes around 25 hours on a GTX-1080 GPU for executing 240K iterations with input image resolution of $256\times512$. The popular Adam optimizer is used with $\beta_1=0.9$, $\beta_2=0.99$ and a learning rate of $1e^{-4}$. The learning rate is reduced by half after ${3/5}^{th}$ and further by half after ${4/5}^{th}$ of total iterations respectively. All the weights $(\lambda^1_{ap},\lambda^2_{ap},\lambda^3_{ap},\lambda_{ds},\lambda^d_{lr})$ are fixed to $1.0$. In order to avoid over smoothening of disparity at final scale, we choose $\lambda_{ds}=0.1/s$ where s is the scale of the disparity with respective to the input resolution. Different kinds of data augmentations are applied to the input images to avoid over-fitting. This includes applying flapping and cropping with a random probability of $0.5$ and color augmentations, such as, random brightness, random gamma and randomly shifting colors in the ranges of $[0.5,2.0],[0.8,1.2],[0.8,1.2]$.  The evaluation of results obtained  is compared and analyzed next in the following sections.

\begin{table*}

\caption{\small Performance Comparison results when evaluated using KITTI split. The testing set consists of 200 images, taken from 28 different scenes. Each image in the test set is associated with the disparity ground truth. The presented results of the state-of-the-art technique \cite{monodepth17} is achieved after implementing their code for the same validation set using our hardware setup. The cells in blue color show the accuracy metric (higher the value, better the performance) and the remaining columns give error metrics (lower value gives better performance).}  
\label{tab:kitti_split_result}
\centering
\scriptsize{
\renewcommand{\arraystretch}{1.1}
\begin{tabular}{ |p{1.9cm}|p{1cm}|p{1cm}|p{1cm}|p{1.3cm}|p{1.2cm}|p{1.2cm}|p{1.2cm}|p{1.2cm}|  }
 \hline
  Method & Abs Rel & Sq Rel & RMSE & logRMSE&D1-all & \cellcolor[HTML]{7BABF7} $\delta <$1.25
  &\cellcolor[HTML]{7BABF7}$\delta <$1.25$^2$ & \cellcolor[HTML]{7BABF7}$\delta <$1.25$^3$ \\
 \hline
 
 Monodepth\cite{monodepth17} & 0.124& 1.388& 6.125  & 0.217 & 30.272 & 0.841     & 0.936 &  0.975\\

 \textbf{Ours Depth}    & 0.1192  &  1.2891 & 5.959 & 0.214 & 30.406 &  0.840  & 0.937 &0.974\\
 
 \textbf{Ours Depth+DIS} & 0.1166  &  1.0929 & 5.751 & 0.209 & 30.633 &  0.841  & 0.939 &0.976\\
		      
 \hline
\end{tabular}}
\end{table*}

\subsection{Depth Evaluation on KITTI DataSet:}

	KITTI 2015 \cite{geiger2013vision} is one of the widely used datasets for benchmarking in this field. It is comprised of 61 different outdoor driving sequences with 42,382 images of resolution $1242\times345$. The dataset is divided into two splits namely, \emph{stereo split} and \emph{eigen split} which are commonly used for evaluating the performance of various deep network models as reported in the literature \cite{monodepth17}\cite{babu2018deeper}\cite{eigen2014depth}.  We also use the same splits for evaluating the performance of our depth estimation model and compare it against the existing state-of-the-art methods in this category. The comparison results for \emph{stereo split} is shown in Table \ref{tab:kitti_split_result} and those for \emph{eigen split} is shown in Table \ref{tab:kitti_eigen_result}. As one can observe, the the proposed UnDEMoN 2.0 model provides the best depth and estimation results for both of these two splits compared to all other state-of-the-art methods. In these tables, `\textbf{Ours D}' refers to the disparity estimation model similar to that used in previous UnDEMoN \cite{babu2018deeper} that only uses a bi-linear image sampler (BIS). The label `\textbf{Ours D+DIS}' refers to the model that uses a deep image sampler (DIS) along with the standard bi-linear image sampler. Finally, the label `\textbf{Ours D+DIS+P}' refers to the proposed UnDEMoN 2.0 model that combines both depth and pose estimation networks.

It should be noted that the most of the existing models \cite{garg2016unsupervised},\cite{babu2018deeper},\cite{monodepth17},\cite{li2018undeepvo} are quite bigger in size. For instance, the models like Monodepth \cite{monodepth17}, SfmLearner\cite{zhou2017unsupervised}, Vid2Depth \cite{mahjourian2018unsupervised} have used U-net \cite{RFB15a} like architecture  which is composed of about 32 Million trainable weights. In contrast, we use a trimmed version of previously reported UnDEMoN architecture \cite{babu2018deeper} that only uses 8 Million parameters while providing best in class performance in this category of algorithms. Quantitatively, UnDEMoN 2.0 provides about 4.47\% improvement over UnDEMoN \cite{babu2018deeper} in terms of RMSE error. This improvement is about 14\%, 8.86\%, 21\% and 18\% in case of Vid2Depth \cite{mahjourian2018unsupervised}, DeepFeat-VO \cite{zhan2018unsupervised} ,SfMLearner \cite{zhou2017unsupervised} and UnDeepVO \cite{li2018undeepvo} respectively. These findings are further corroborated by analysing the qualitative comparison provided in Figure \ref{fig:depth_results} where one can notice that the performance of UnDEMoN and UnDEMoN 2.0 is visually similar but better than the other methods. These results clearly demonstrates that the inclusion of deep image sampler significantly improves the performance of depth estimation models that only rely on bi-linear sampler for image reconstruction. This can be easily verified by analysing the figure \ref{fig:reconst_loss} which shows that a combination of DISNet and BIS provides lower image reconstruction losses compared to those obtained using only Bi-linear sampler (BIS).  
\begin{table*}
\caption{\small Absolute Trajectory Error (ATE) \cite{mur2015orb} for Translation and Rotation on KITTI eigen split dataset averaged over all 3-frame snippets (lower is better). As one can see, UnDEMoN 2.0 outperforms the monocular versions UnDEMoN \cite{babu2018deeper}, SfMLearner\_noPP \cite{zhou2017unsupervised} and VISO\_M \cite{Geiger2011IV} and is comparable with SfMLearner\_PP \cite{zhou2017unsupervised} and VISO\_S \cite{Geiger2011IV} (stereo version of VISO). Here, the terms $t_{ate}$ and $r_{ate}$ stand for translational absolute trajectory error and rotational absolute trajectory error respectively. }
\begin{center}
\scriptsize{
\label{tab:ate}
\begin{tabular}{|c|cc|cc|cc|cc|cc|cc|} \toprule[1.5pt]
 \head{Seq} & \multicolumn{2}{c|}{UnDEMoN 2.0} & \multicolumn{2}{c|}{UnDEMoN\cite{babu2018deeper}} & \multicolumn{2}{c|}{SfMLearner\_noPP \cite{zhou2017unsupervised}} & 
    \multicolumn{2}{c|}{SfMLearner\_PP \cite{zhou2017unsupervised}} & \multicolumn{2}{c|}{VISO2\_S \cite{Geiger2011IV}} & \multicolumn{2}{c|}{VISO\_M \cite{Geiger2011IV}}\\
  & $t_{ate}$ &  $r_{ate}$ & $t_{ate}$ &  $r_{ate}$ &  $t_{ate}$  &   $r_{ate}$ & $t_{ate}$ &   $r_{ate}$
& $t_{ate}$ &   $r_{ate}$   \\ \hline 
  00 &  0.0607 & 0.0014& 0.0644 & 0.0013 & 0.7366 & 0.0040 & 0.0479 & 0.0044 & 0.0429 & 0.0006 & 0.1747 & 0.0009\\
  04 &  0.0690 & 0.0007  &0.0974 & 0.0008  & 1.5521 & 0.0027 & 0.0913  & 0.0027  & 0.0949 & 0.0010 & 0.2184 & 0.0009\\
    05 &  0.0659  & 0.0009 &0.0696  & 0.0009 & 0.7260 & 0.0036 & 0.0392  & 0.0036  & 0.0470 & 0.0004 & 0.3787 & 0.0013\\
     07 &  0.0730 & 0.0011& 0.0742 & 0.0011 & 0.5255 & 0.0036 & 0.0345 & 0.0036  & 0.0393 & 0.0004 & 0.4803 & 0.0018\\
  \bottomrule[1.5pt]
\end{tabular}}
\end{center}
\end{table*}
\subsection{Pose Evaluation on KITTI DataSet}
The performance of Pose Net is evaluated using the image sequences of the Odometry split which are in the test set of the eigen split of the KITTI dataset as explained in \cite{babu2018deeper}. We use Absolute Trajectory Error \cite{mur2015orb} as a measure for comparing the performance of our model with other state-of-the-art methods in the field. The resulting comparison is shown in  Table \ref{tab:ate}. The SfMLearner \cite{zhou2017unsupervised} employs a post processing stage that uses ground truth pose to obtain the absolute scale information and is referred to by using the suffix \_PP. For a fair comparison with our method that does not use any ground truth, we obtain the results for SfMLearner by removing this post processing step and is denoted by the suffix \_noPP. Similarly, we compare the performance of our algorithm with the monocular (VISO\_M) and stereo (VISO\_S) version of the VISO \cite{Geiger2011IV} model which is a known traditional method in this category. As one can see the UnDEMoN 2.0 outperforms UnDEMoN, SfMLearner\_noPP and VISO\_M and is comparable to the VISO\_S and SfMLearner\_PP that use ground truth information explicitly.  

\section{Conclusion} \label{sec:conc}
This paper proposes several modifications to the previously published UnDEMoN architecture to improve its performance in estimating depth and pose directly from monocular images. The improvement is achieved by including a deep network based image sampler into the UnDEMoN architecture. This combined with a traditional bi-linear sampler provides superior image reconstruction capabilities which, in turn, improves the depth and pose estimation accuracy. This is motivated by the observation that a deep network provides higher order regression and larger input search space compared to a bi-linear sampler that uses a quadratic regression and only four points in the input space to estimate a pixel in the target image. In addition, the geometry constraints obtained from a bi-linear sampler helps in reducing the solution space of an ill-posed problem of this kind. The resulting network is named UnDEMoN 2.0 and  is shown to outperform all existing deep network models that use unsupervised learning for depth and pose estimation. This feat is achieved by using one of the lightest deep learning models that uses only a quarter of the total number of tunable parameters used in UnDEMoN. However, the proposed model considers the scene to be static and hence, can not deal with moving objects. This can be addressed by incorporating optical flow into the deep model as suggested in several other works, e.g. \cite{yin2018geonet}. Similarly, the reconstructed image quality can be further improved by employing a discriminative network trained in an adversarial fashion. A future direction for this work would be to incorporate these two concepts to achieve complete understanding of scene that includes both static as well as moving objects.    
\section{Appendix}

\subsection{Model Architecture}
This section provides an elaborate description of the proposed UnDEMoN 2.0 architecture. The architecture comprises of three networks: DispNet, PoseNet and DISNet. DispNet predicts disparity from a monocular image, PoseNet estimates ego-motion using a sequence of monocular images and finally DISNet uses the input image and the predicted disparity to reconstruct the opposite stereo view.

\subsubsection{DispNet}
DispNet predicts left to right and right to left disparities for a given monocular image as an input. It is further used to calculate depth at every pixel.  Let us consider $I_l$ as the input image and the network predictions as $d^l$ and $d^r$ for left to right and right to left disparities respectively. The pixel locations of the right image $I'_r$ is can be calculated as $I_l+d^r$. We have used inverse image warping technique to add color intensities to the new pixels of $I'_r$ by sampling from left image using bi-linear interpolation. Once the disparity is known the depth $D$ can be calculated as $D=\frac{b\times f}{d}$ where $b$ is the baseline distance between the stereo cameras and $f$ is the rectified focal length. The network has the ability of multi-scale disparity estimation which in turn enables the network to avoid gradient locality problem. An architectural overview of the network is provided in the Table:\ref{tab:dispNet}.  Relu is used as a non-linear activation function in all the layers of the network expect the output layers (disp1,disp2,disp3, disp4). We restrict the the maximum disparity to be 30\% of the input image width by employing a sigmoid activation function in those layers. 
\begin{table*}[!thpb]
\centering
\caption{DispNet architecture, where \textbf{channels} represents input/output channels for the each layer and \textbf{in}, \textbf{out} represents the input/output resolution scale with respective to the input image resolution. Encoder and Decoder parts are shown differently as two sections. In decoder \textbf{*} represents $2\times$ upscaling of its current scale.  }
\begin{tabular}
{ |c|c|c|c|c|c|c| } 
\hline
 Layer  &  Kernel    &  Stride & channels   & in & out & input\\
\hline
{conv1}  & $7\times7$ &   2    &   $3/32$    & 1  &  2  & Left \\ 
{conv1b} & $7\times7$ &   1    &   $32/32$    & 2  &  2  & conv1 \\ 

{conv2}  & $5\times5$ &   2    &   $32/64$    & 2  &  4  & conv1b \\ 
{conv2b} & $5\times5$ &   1    &   $64/64$    & 4  &  4  & conv2 \\ 

{conv3}  & $3\times3$ &   2    &   $64/128$   & 4  &  8  & conv2b \\ 
{conv3b} & $3\times3$ &   1    &   $128/128$   & 8  &  8  & conv3 \\ 

{conv4}  & $3\times3$ &   2    &   $128/256$   & 8  & 16  & conv3b \\ 
{conv4b} & $3\times3$ &   1    &   $256/256$   & 16 & 16  & conv4 \\ 

{conv5}  & $3\times3$ &   2    &   $256/512$   & 16  &  32  & conv4b \\ 
{conv5b} & $3\times3$ &   1    &   $512/512$   & 32  &  32  & conv5 \\ 

\hline
\hline

{upconv5} & $3\times3$ &   2    &   $512/256$    & 32  &  16  & conv5b \\ 
{iconv5}  & $3\times3$ &   1    &   $512/256$    & 16  &  16  & upconv5+conv4b \\ 

{upconv4} & $3\times3$ &   2    &   $256/128$    & 16  &  8  & inconv5 \\ 
{iconv4}  & $3\times3$ &   1    &   $256/128$    & 8  &  8   & upconv4+conv3b \\

\textbf{disp4}  & $3\times3$ &   1    &$128/2$    & 8  &  8   & iconv4\\
 
{upconv3} & $3\times3$ &   2    &   $128/64$     & 8  &  4 & iconv4 \\ 
{iconv3}  & $3\times3$ &   1    &   $130/64$     & 4  &  4  & upconv3+conv2b+disp4* \\ 
\textbf{disp3}  
         & $3\times3$ &   1      &$64/2$        & 4  &  4   & iconv3\\

{upconv2} & $3\times3$ &   2    &   $64/32$      & 4  &  3  & iconv3 \\ 
{iconv2}  & $3\times3$ &   1    &   $66/32$      & 2  &  2  & upconv2+conv1b+disp3* \\

\textbf{disp2}  
         & $3\times3$ &   1    &    $32/2$      & 2  &  2   & iconv2\\

{upconv1} & $3\times3$ &   2    &   $32/16$    & 2  &  1  & iconv2 \\ 
{iconv1}  & $3\times3$ &   1    &   $18/16$    & 1  &  1  & upconv1+disp2* \\ 

\textbf{disp1}   
         & $3\times3$ &   1    &   $16/2$     & 1  &  1   & iconv1\\ 
         
\hline

\end{tabular}
\label{tab:dispNet}
\end{table*}

\subsubsection{PoseNet }

The detailed network architecture of the PoseNet is shown in Table:\ref{tab:poseNet}. Relu  is used as an activation function in all the layers except in $r, t$ layers. No activation functions used for the $r, t$ layers. The network take sequence of monocular images (n-frame snippets) and estimates $(n-1)\times 6$ pose vectors. Lets consider two temporally aligned images as $I_t,I_{t+1}$ and the predicted pose vectors are given by $P_{t\rightarrow t+1}$. Having known these parameters, one can easily reconstruct the image $I'_t$ from the forward image $I_{t+1}$ using the $P, D_t, K$ by inverse image warping. Where $D_t$ is the depth of the image $I_t$ which is predicted using DispNet and K is the calibration matrix. Mathematically, this can be represented as 

\begin{equation}
\label{eq:view_synthesis}
p^{t+1}_i = K P_{t\rightarrow t+1} {D}^t_iK^{-1} p^t_i
\end{equation}

\begin{table*}[!thpb]
\centering
\caption{PoseNet architecture, where \textbf{channels} are defined as input/output channels for the each layer and \textbf{in}, \textbf{out} represents the input/output resolution scale with respective to the input image resolution. PoseNet is composed of convolutional encoder followed by Fully connected layers. Fully connected layers are used in $FC, t, r$ layers. $H,W$ are the input image height and width respectively and n is the no of images in the input sequence.}
\begin{tabular}
{ |c|c|c|c|c|c|c| } 
\hline
 Layer  &  Kernel    &  Stride & channels   & in & out & input\\
\hline
{conv1}  & $7\times7$ &   2    &   $n\times3/16$    & 1  &  2  & Left \\ 

{conv2}  & $5\times5$ &   2    &   $16/32$    & 2  &  4  & conv1\\ 

{conv3}  & $3\times3$ &   2    &   $32/64$   & 4  &  8  & conv2\\ 

{conv4}  & $3\times3$ &   2    &   $64/128$   & 8  & 16  & conv3 \\ 

{conv5}  & $3\times3$ &   2    &   $128/256$   & 16  &  32  & conv4 \\ 

{conv6}  & $3\times3$ &   2    &   $256/256$   & 32  &  64  & conv5 \\ 

{conv7}  & $3\times3$ &   2    &   $256/512$   & 64  &  128  & conv6 \\ 
\hline
FC$_1$     &			 &        &   $\frac{H}{128}\times\frac{W}{128}\times512/512$      &      &   &conv7$_{flat}$    \\

$t$     &            &        &   $512/(n-1)\times 3$ &  &  & FC$_1$\\

$r$     &            &        &   $512/(n-1)\times 3$ &  &  & FC$_1$\\

\hline

\end{tabular}
\label{tab:poseNet}
\end{table*}

\subsubsection{DISNet}

The proposed Deep Image Sampling network (DISNet) is also similar to the DispNet except the output layers.  The output layers have 3-channels for $R,G$ $B$ respectively. However, these layers have no activation functions. The network takes a left or a right image ($I_l\,\text{or}\,I_r$) and the corresponding disparity $(d^r\,\text{or}\, d^l)$ as input to predict the opposite stereo image ($I''_r\,\text{or}\,I''_l$). The DISNet also features multi scale image estimations.

\subsection{Misc:}
This section gives the details of the matrices used to analysis and compare the results with the existing state-of-the-art methods. Some more depth estimation results are also provided later in this section where the performances of the proposed UNDEMON 2.0 is shown to provide superior results over the current state-of-the art.

\begin{figure*}
\centering
\includegraphics[scale=0.35]{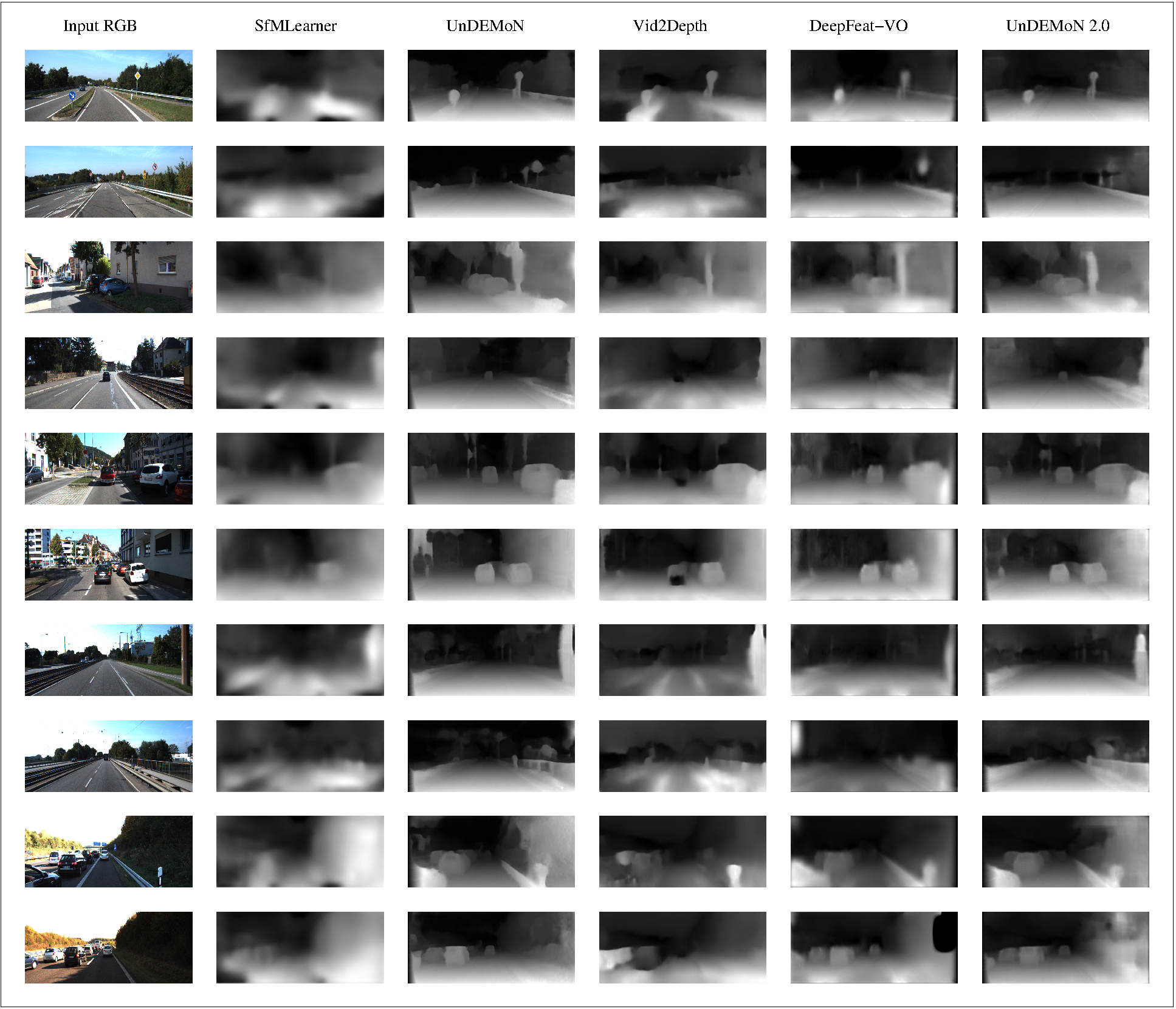}
\caption{\small Additional qualitative results of UnDEMoN 2.0 on KITTI Eigne Split with various state of the art methods.}
\label{fig:depth_results}
\end{figure*}

\subsubsection{Structural Similarity Index:}

Structural similarity index is a perceptual metric to estimate how much the reconstructed image degrades with reference to the original image. It is composed of three terms, namely luminance mask, contrast mask and a structural term. The index is given as follows. 

\begin{equation}
SSIM(I,\tilde{I})=\frac{(2\mu_I\mu_{\tilde{I}}+c_1)(2\sigma_{I\tilde{I}}+c_2)}{(\mu^2_{I}+\mu^2_{\tilde{I}}+c_1)(\sigma^2_I+\sigma^2_{\tilde{I}})}
\end{equation}
where, the values of $c_1$, $c_2$ are set to $0.01^2,0.03^2$ respectively. 
In all the three losses presented in the paper we  have used a window size of $3\times3$. 
\subsubsection{Evaluation Metrics}
We have used several standard evaluation metrics to show the efficacy of the proposed approach. These are given in the following. Lets assume that the input image has a total $N$ no of pixels. For a pixel $i$, the predicted value is $y_i$ and the ground truth is $y^*_i$. Mathematically the metrics can be represented as follows, 

\begin{eqnarray}\nonumber
\text{Abs Relative difference:}\, \frac{1}{N}\sum_{i\epsilon N}\frac{|y_i-y_i^*|}{y_i^*} \\\nonumber
\end{eqnarray}
\begin{eqnarray}\nonumber
\text{Squared relative difference:}\,\frac{1}{N}\sum_{i\epsilon N}{\frac{||y_i-y_i^*||^2}{y_i^*}}\\\nonumber
\text{RMSE Linear:}\,\sqrt[.]{\frac{1}{N}\sum_{i\epsilon N}{||y_i-y_i^*||^2}}\\\nonumber
\text{RMSE log:}\,\sqrt[.]{\frac{1}{N}\sum_{i\epsilon N}{||\text{log}\,y_i-\text{log}\,y_i^*||^2}}\\\nonumber
\text{Threshold: \% of $y_i$ s.t max($\frac{y_i}{y_i^*},\frac{y_i^*}{y_i}$)}=\delta< \text{thr}\\\nonumber
\text{thr}\,\epsilon\,[1.25,1.25^2,1.25^3]\\\nonumber
\end{eqnarray}
\subsubsection{More Qualitative Results:}
We have shown more qualitative results of UnDEMoN 2.0 against several state of the art methods in Fig.\ref{fig:depth_results} for KITTI dataset using Eigen split.

\bibliographystyle{abbrv}

\begin{thebibliography}{10}

\bibitem{abadi2016tensorflow}
M.~Abadi, P.~Barham, J.~Chen, Z.~Chen, A.~Davis, J.~Dean, M.~Devin,
  S.~Ghemawat, G.~Irving, M.~Isard, et~al.
\newblock Tensorflow: a system for large-scale machine learning.
\newblock In {\em OSDI}, volume~16, pages 265--283, 2016.

\bibitem{babu2018deeper}
M.~BabuV, A.~Majumder, K.~Das, S.~Kumar, et~al.
\newblock Undemon: Unsupervised deep network for depth and ego-motion
  estimation.
\newblock {\em Proceedings of the IEEE/RSJ International Conference on
  Intelligent Robots}, 2018.

\bibitem{eigen2014depth}
D.~Eigen, C.~Puhrsch, and R.~Fergus.
\newblock Depth map prediction from a single image using a multi-scale deep
  network.
\newblock In {\em Advances in neural information processing systems}, pages
  2366--2374, 2014.

\bibitem{furukawa2015multi}
Y.~Furukawa, C.~Hern{\'a}ndez, et~al.
\newblock Multi-view stereo: A tutorial.
\newblock {\em Foundations and Trends{\textregistered} in Computer Graphics and
  Vision}, 9(1-2):1--148, 2015.

\bibitem{garg2016unsupervised}
R.~Garg, V.~K. BG, G.~Carneiro, and I.~Reid.
\newblock Unsupervised cnn for single view depth estimation: Geometry to the
  rescue.
\newblock In {\em European Conference on Computer Vision}, pages 740--756.
  Springer, 2016.

\bibitem{geiger2013vision}
A.~Geiger, P.~Lenz, C.~Stiller, and R.~Urtasun.
\newblock Vision meets robotics: The kitti dataset.
\newblock {\em The International Journal of Robotics Research},
  32(11):1231--1237, 2013.

\bibitem{stereoscan}
A.~Geiger, J.~Ziegler, and C.~Stiller.
\newblock Stereoscan: Dense 3d reconstruction in real-time.
\newblock In {\em Intelligent Vehicles Symposium (IV), 2011 IEEE}, pages
  963--968. Ieee, 2011.

\bibitem{Geiger2011IV}
A.~Geiger, J.~Ziegler, and C.~Stiller.
\newblock Stereoscan: Dense 3d reconstruction in real-time.
\newblock In {\em Intelligent Vehicles Symposium (IV)}, 2011.

\bibitem{monodepth17}
C.~Godard, O.~{Mac Aodha}, and G.~J. Brostow.
\newblock Unsupervised monocular depth estimation with left-right consistency.
\newblock In {\em CVPR}, 2017.

\bibitem{handa2014benchmark}
A.~Handa, T.~Whelan, J.~McDonald, and A.~J. Davison.
\newblock A benchmark for rgb-d visual odometry, 3d reconstruction and slam.
\newblock In {\em Robotics and automation (ICRA), 2014 IEEE international
  conference on}, pages 1524--1531. IEEE, 2014.

\bibitem{kendall2015posenet}
A.~Kendall, M.~Grimes, and R.~Cipolla.
\newblock Posenet: A convolutional network for real-time 6-dof camera
  relocalization.
\newblock In {\em IEEE International Conference on Computer Vision (ICCV),
  2015}, pages 2938--2946. IEEE, 2015.

\bibitem{kuznietsov2017semi}
Y.~Kuznietsov, J.~St{\"u}ckler, and B.~Leibe.
\newblock Semi-supervised deep learning for monocular depth map prediction.
\newblock In {\em Proc. of the IEEE Conference on Computer Vision and Pattern
  Recognition}, pages 6647--6655, 2017.

\bibitem{li2018undeepvo}
R.~Li, S.~Wang, Z.~Long, and D.~Gu.
\newblock Undeepvo: Monocular visual odometry through unsupervised deep
  learning.
\newblock In {\em 2018 IEEE International Conference on Robotics and Automation
  (ICRA)}, pages 7286--7291. IEEE, 2018.

\bibitem{liu2015deep}
F.~Liu, C.~Shen, and G.~Lin.
\newblock Deep convolutional neural fields for depth estimation from a single
  image.
\newblock In {\em Proceedings of the IEEE Conference on Computer Vision and
  Pattern Recognition}, pages 5162--5170, 2015.

\bibitem{learning_depth_mono_cnf}
F.~Liu, C.~Shen, G.~Lin, and I.~Reid.
\newblock Learning depth from single monocular images using deep convolutional
  neural fields.
\newblock {\em IEEE transactions on pattern analysis and machine intelligence},
  38(10):2024--2039, 2016.

\bibitem{luo2018single}
Y.~Luo, J.~Ren, M.~Lin, J.~Pang, W.~Sun, H.~Li, and L.~Lin.
\newblock Single view stereo matching.
\newblock In {\em Proceedings of the IEEE Conference on Computer Vision and
  Pattern Recognition}, pages 155--163, 2018.

\bibitem{mahjourian2018unsupervised}
R.~Mahjourian, M.~Wicke, and A.~Angelova.
\newblock Unsupervised learning of depth and ego-motion from monocular video
  using 3d geometric constraints.
\newblock In {\em Proceedings of the IEEE Conference on Computer Vision and
  Pattern Recognition}, pages 5667--5675, 2018.

\bibitem{mahmood2018deep}
F.~Mahmood and N.~J. Durr.
\newblock Deep learning-based depth estimation from a synthetic endoscopy image
  training set.
\newblock In {\em Medical Imaging 2018: Image Processing}, volume 10574, page
  1057421. International Society for Optics and Photonics, 2018.

\bibitem{pose_ar}
E.~Marchand, H.~Uchiyama, and F.~Spindler.
\newblock Pose estimation for augmented reality: a hands-on survey.
\newblock {\em IEEE transactions on visualization and computer graphics},
  22(12):2633--2651, 2016.

\bibitem{disp_net}
N.~Mayer, E.~Ilg, P.~Hausser, P.~Fischer, D.~Cremers, A.~Dosovitskiy, and
  T.~Brox.
\newblock A large dataset to train convolutional networks for disparity,
  optical flow, and scene flow estimation.
\newblock In {\em Proceedings of the IEEE Conference on Computer Vision and
  Pattern Recognition}, pages 4040--4048, 2016.

\bibitem{mur2015orb}
R.~Mur-Artal, J.~M.~M. Montiel, and J.~D. Tardos.
\newblock Orb-slam: a versatile and accurate monocular slam system.
\newblock {\em IEEE Transactions on Robotics}, 31(5):1147--1163, 2015.

\bibitem{RFB15a}
O.~Ronneberger, P.Fischer, and T.~Brox.
\newblock U-net: Convolutional networks for biomedical image segmentation.
\newblock In {\em Medical Image Computing and Computer-Assisted Intervention
  (MICCAI)}, volume 9351 of {\em LNCS}, pages 234--241. Springer, 2015.
\newblock (available on arXiv:1505.04597 [cs.CV]).

\bibitem{scharstein2002taxonomy}
D.~Scharstein and R.~Szeliski.
\newblock A taxonomy and evaluation of dense two-frame stereo correspondence
  algorithms.
\newblock {\em International journal of computer vision}, 47(1-3):7--42, 2002.

\bibitem{charbonnier_loss}
D.~Sun, S.~Roth, and M.~J. Black.
\newblock A quantitative analysis of current practices in optical flow
  estimation and the principles behind them.
\newblock {\em International Journal of Computer Vision}, 106(2):115--137,
  2014.

\bibitem{taylor2016efficient}
J.~Taylor, L.~Bordeaux, T.~Cashman, B.~Corish, C.~Keskin, T.~Sharp, E.~Soto,
  D.~Sweeney, J.~Valentin, B.~Luff, et~al.
\newblock Efficient and precise interactive hand tracking through joint,
  continuous optimization of pose and correspondences.
\newblock {\em ACM Transactions on Graphics (TOG)}, 35(4):143, 2016.

\bibitem{demon}
B.~Ummenhofer, H.~Zhou, J.~Uhrig, N.~Mayer, E.~Ilg, A.~Dosovitskiy, and
  T.~Brox.
\newblock Demon: Depth and motion network for learning monocular stereo.
\newblock In {\em IEEE Conference on Computer Vision and Pattern Recognition
  (CVPR)}, volume~5, 2017.

\bibitem{deepvo}
S.~Wang, R.~Clark, H.~Wen, and N.~Trigoni.
\newblock Deepvo: Towards end-to-end visual odometry with deep recurrent
  convolutional neural networks.
\newblock In {\em Robotics and Automation (ICRA), 2017 IEEE International
  Conference on}, pages 2043--2050. IEEE, 2017.

\bibitem{wang2018occlusion}
Y.~Wang, Y.~Yang, Z.~Yang, L.~Zhao, and W.~Xu.
\newblock Occlusion aware unsupervised learning of optical flow.
\newblock In {\em Proceedings of the IEEE Conference on Computer Vision and
  Pattern Recognition}, pages 4884--4893, 2018.

\bibitem{wang2004image}
Z.~Wang, A.~C. Bovik, H.~R. Sheikh, and E.~P. Simoncelli.
\newblock Image quality assessment: from error visibility to structural
  similarity.
\newblock {\em IEEE transactions on image processing}, 13(4):600--612, 2004.

\bibitem{yin2018geonet}
Z.~Yin and J.~Shi.
\newblock Geonet: Unsupervised learning of dense depth, optical flow and camera
  pose.
\newblock In {\em CVPR}, 2018.

\bibitem{zhan2018unsupervised}
H.~Zhan, R.~Garg, C.~S. Weerasekera, K.~Li, H.~Agarwal, and I.~Reid.
\newblock Unsupervised learning of monocular depth estimation and visual
  odometry with deep feature reconstruction.
\newblock In {\em Proceedings of the IEEE Conference on Computer Vision and
  Pattern Recognition}, pages 340--349, 2018.

\bibitem{zhou2017unsupervised}
T.~Zhou, M.~Brown, N.~Snavely, and D.~G. Lowe.
\newblock Unsupervised learning of depth and ego-motion from video.
\newblock In {\em CVPR}, 2017.
\end{thebibliography}

\end{document}